\documentclass[twocolumn, 10pt, letterpaper]{article}

\usepackage{amssymb}
\usepackage{amsmath}
\usepackage[colorlinks, bookmarks = false]{hyperref}
\usepackage{graphicx}
\usepackage{tabularx}
\usepackage{multirow}
\usepackage{adjustbox}
\usepackage{url}
\usepackage{style/cvpr}
\usepackage{times}

\cvprfinalcopy

\title{\bf EfficientPose: An efficient, accurate and scalable end-to-end 6D multi object pose estimation approach}

\author{Yannick Bukschat\\
				\small{Steinbeis Transferzentrum an der Hochschule Mannheim}\\
				\texttt{\small yannick.bukschat@stw.de}
				\and
				Marcus Vetter\\
				\small{ESM-Institut, Hochschule Mannheim}\\
				\texttt{\small m.vetter@hs-mannheim.de}
}

\begin{document}

\maketitle

\begin{abstract}
\textit{In this paper we introduce EfficientPose, a new approach for 6D object pose estimation. Our method is highly accurate, efficient and scalable over a wide range of computational resources. Moreover, it can detect the 2D bounding box of multiple objects and instances as well as estimate their full 6D poses in a single shot. This eliminates the significant increase in runtime when dealing with multiple objects other approaches suffer from. These approaches aim to first detect 2D targets, e.g. keypoints, and solve a Perspective-n-Point problem for their 6D pose for each object afterwards. We also propose a novel augmentation method for direct 6D pose estimation approaches to improve performance and generalization, called 6D augmentation. Our approach achieves a new state-of-the-art accuracy of \textbf{97.35\%} in terms of the ADD(-S) metric on the widely-used 6D pose estimation benchmark dataset Linemod using RGB input, while still running \textbf{end-to-end at over 27 FPS}. Through the inherent handling of multiple objects and instances and the fused single shot 2D object detection as well as 6D pose estimation, our approach runs even with \textbf{multiple objects (eight) end-to-end at over 26 FPS}, making it highly attractive to many real world scenarios. Code will be made publicly available at {\small \url{https://github.com/ybkscht/EfficientPose}}.}
\end{abstract}

\section{Introduction}
\label{section_introduction}

\begin{figure}
\centering
\includegraphics[width = \linewidth]{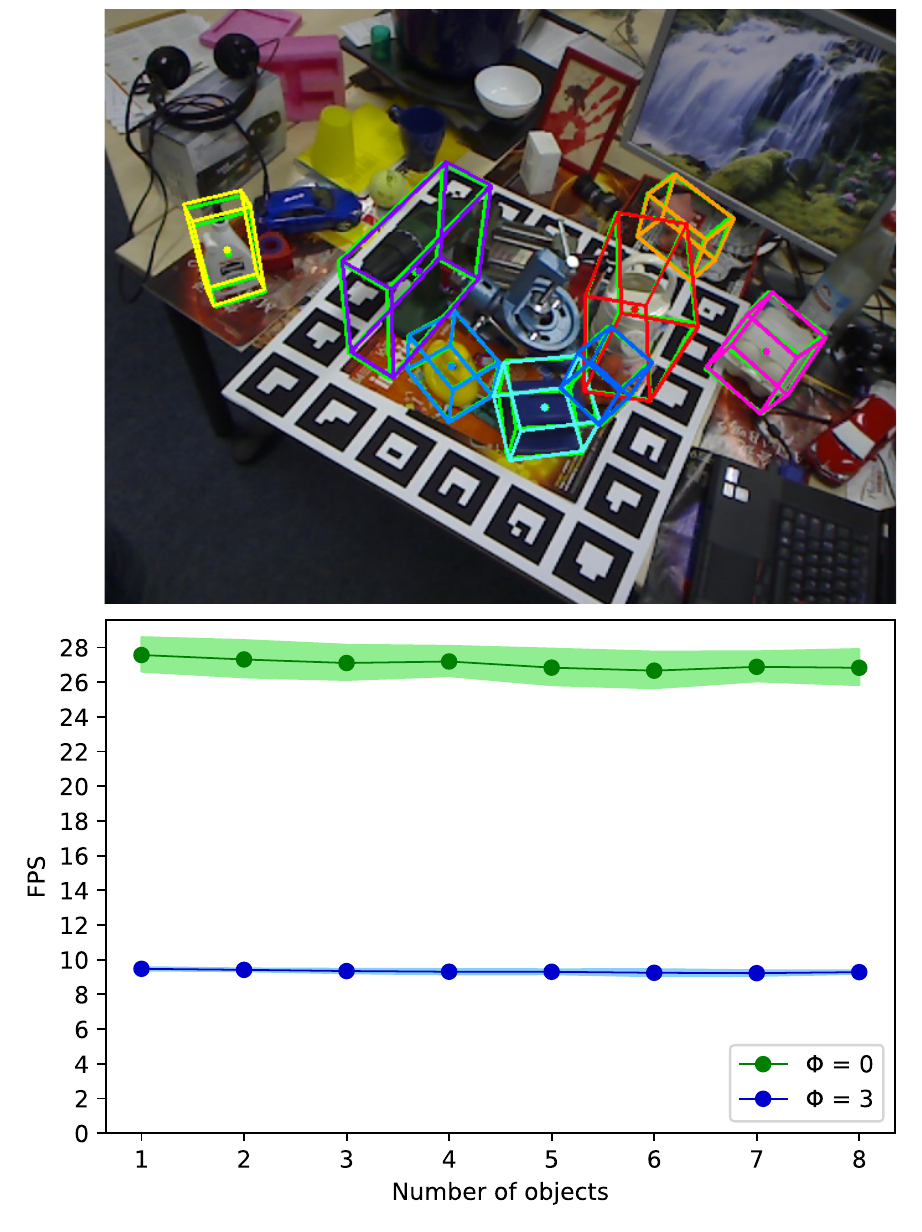}
\caption{\textbf{Top:} Example prediction for qualitative evaluation of our $\phi = 0$ model performing single shot 6D multi object pose estimation on the Occlusion test set while running end-to-end at over 26 FPS. Green 3D bounding boxes visualize ground truth poses while our estimated poses are represented by the other colors.\\
\textbf{Bottom:} Average end-to-end runtimes in FPS of our $\phi = 0$ and $\phi = 3$ model on the Occlusion test set \wrt the number of objects per image. Shaded areas represent the standard deviations.}
\label{figure_title}
\end{figure}

Detecting objects of interest in images is an important task in computer vision and a lot of works in this research field developed highly accurate methods to tackle this problem \cite{DetectorRS}\cite{SpineNet}\cite{AmoebaNet_NAS_FPN}\cite{RetinaNet}\cite{faster_rcnn}. More recently some works not only focused on the accuracy but also on the efficiency to make their methods applicable in real world scenarios with computational and runtime limitations\cite{CSPNet}\cite{EfficientDet}. For example Tan \etal\cite{EfficientDet} developed a highly scalable and efficient approach, called EfficientDet, that can easily be scaled over a high range of computational resources, speed and accuracy, with a single hyperparameter. But for some tasks like robotic manipulation, autonomous vehicles and augmented reality, it is not enough to detect only the 2D bounding boxes of the objects in an image, but to also estimate their 6D poses. Most of the recent works achieving state-of-the-art accuracy in the field of 6D object pose estimation with RGB input rely on an approach that detects 2D targets, e.g. keypoints, of the objects of interest in the image first and solve for their 6D poses with a PnP-algorithm afterwards \cite{YOLO6D}\cite{Pix2Pose}\cite{PVNet}\cite{DPOD}\cite{CDPN}\cite{HybridPose}. While they achieve good 6D pose estimation accuracy and since some of them are also relatively fast in terms of single object pose estimation, the runtime linearly increases with the number of objects. This results from the need to compute the 6D pose via PnP for each object individually. Furthermore, some approaches use a pixel-wise RANSAC-based\cite{RANSAC} voting scheme to detect the needed keypoints, which also has to be performed for each object separately and therefore can be very time consuming \cite{PVNet}\cite{HybridPose}. Moreover, some methods need a separate 2D object detector first to localize and crop the bounding boxes of the objects of interest. These cropped image patches subsequently serve as the input of the actual 6D pose estimation approach which means that the whole method needs to be applied for each detected object separately \cite{Pix2Pose}\cite{CDPN}. For these reasons, those approaches are often not well suited for use cases with multiple objects and runtime limitations, which inhibit their deployment in many real world scenarios.\\
In this work we propose a new approach which does not encounter these issues and still achieves state-of-the-art performance using RGB input on the widely-used benchmark dataset Linemod \cite{Linemod}.
To achieve this, we extend the state-of-the-art 2D object detection architecture family EfficientDets in an intuitive way to also predict the 6D poses of objects. Therefore, we add two extra subnetworks to predict the translation and rotation of objects, analogous to the classification and bounding box regression subnetworks. Since these subnets are relatively small and share the computation of the input feature maps with the already existing networks, we are able to get the full 6D pose very inexpensive without much additional computational cost.
Through the seamless integration in the EfficientDet architecture, our approach is also capable of detecting multiple object categories as well as multiple object instances and can estimate their 6D poses - all within a single shot. Because we regress the 6D pose directly, we need no further post-processing steps like RANSAC and PnP. This makes the runtime of our method nearly independent from the number of objects per image.\\
A key element for our reported state-of-the-art accuracy, in terms of the ADD(-S) metric on the Linemod dataset, turned out to be our proposed 6D augmentation which boosts the performance of our approach enormously. This proposed augmentation technique allows direct 6D pose estimation methods like ours, to also use image rotation and scaling which otherwise would lead to a mismatch between image and annotated poses. Such image manipulations can help to significantly improve performance and generalization when dealing with small datasets like Linemod \cite{RandAugment}\cite{AmoebaNet_NAS_FPN}. 2D+PnP approaches are able to exploit those methods without much effort because the 2D targets can be relatively easy transformed accordingly to the image transformation. Using our proposed augmentation method can help to compensate for that previous advantage of 2D+PnP approaches which arguably could be a reason for the current dominance of those approaches in the field of 6D object pose estimation with RGB input \cite{PVNet}\cite{DPOD}\cite{HybridPose}.\\
Just like the original EfficientDets, our approach is also highly scalable via a single hyperparameter $\phi$ to adjust the network to a wide range of computational resources, speed and accuracy.
Last but not least, because our method needs no further post-processing steps, as already mentioned, and as it is based on an architecture that inherently handles multiple object categories and instances, our approach is relatively easy to use and therefore makes it attractive for many real world scenarios.\\
To sum it all up, our main contributions in this work are as follows:
\begin{itemize}
	\item 6D Augmentation for direct 6D pose estimation approaches to improve performance and generalization, especially when dealing with small datasets.
	\item Extending the state-of-the-art 2D object detection family of EfficientDets with the additional ability of 6D object pose estimation while keeping their advantages like inherent single shot multi object and instance detection, high accuracy, scalability, efficiency and ease of use.
\end{itemize}

\section{Related Work}
\label{section_related_work}
In this section we briefly summarize already existing works that are related to our topic. The deep learning based approaches in the research field of 6D pose estimation using RGB input can mostly be assigned to one of the following two categories - estimating the 6D pose directly or first detecting 2D targets in the given image and then solving a Perspective-n-Point (PnP) problem for the 6D pose. As our method is based on a 2D object detector, we also shortly summarize related work of this research field.

\subsection{Direct estimation of the 6D pose}
\label{subsection_direct_6d_pose_regression}
Probably the most straight forward way to estimate an object's 6D pose is to directly regress it. PoseCNN \cite{PoseCNN} follows this strategy as they internally decouple the translation and rotation estimation parts. They also propose a novel loss function to handle symmetric objects since, due to their ambiguities, the network can be penalized unnecessarily during training when not taking their symmetry into account. This loss function is called ShapeMatch-Loss and we base our own loss, described in \autoref{subsection_transformation_loss}, on that function.\\

Another possibility is to discretize the continuous rotation space into bins and classify them. Kehl \etal\cite{SSD6D} and Sundermeyer \etal\cite{AAE} are using this approach. SSD-6D\cite{SSD6D} extends the 2D object detector SSD\cite{SSD} with that ability while AAE\cite{AAE} aims for learning an implicit rotation representation via auto encoders and assign that estimated rotation to a similar rotation vector in a codebook. However, due to the nature of the discretization process, the so obtained poses are very course and have to be further refined in order to get a relatively accurate 6D pose.

\subsection{2D Detection and PnP}
\label{subsection_2d_detection_and_pnp}
More recently the state-of-the-art accuracy regime of 6D object pose estimation using RGB input only is dominated by approaches that first detect 2D targets of the object in the given image and subsequently solve a Perspective-n-Point problem for their 6D pose \cite{PVNet}\cite{HybridPose}\cite{DPOD}\cite{CDPN}\cite{Pix2Pose}\cite{BPnP}. This approach can be further split in two categories - keypoint-based \cite{PVNet}\cite{HybridPose}\cite{BPnP}\cite{BB8}\cite{DOPE}\cite{YOLO6D} and dense 2D-3D correspondence methods \cite{DPOD}\cite{CDPN}\cite{Pix2Pose}. The keypoint-based methods predict either the eight 2D projections of the cuboid corners of the 3D model as keypoints \cite{BB8}\cite{DOPE}\cite{YOLO6D} or choose keypoints on the object's surface, often selected with the farthest point sampling algorithm \cite{PVNet}\cite{HybridPose}\cite{BPnP}. Since the cuboid corners are often not on the object's surface, those keypoints are usually harder to predict than their surface counterparts, but instead only need the 3D cuboid of the object and not the complete 3D model. Because keypoints can also be invisible in the image due to occlusion or truncation, some methods perform a pixel-wise voting scheme where each pixel of the object predicts a vector pointing to the keypoint \cite{PVNet}\cite{HybridPose}. The final keypoints are estimated using RANSAC\cite{RANSAC}, which makes it more robust to outliers when dealing with occlusion.\\

The dense 2D-3D correspondence methods predict the corresponding 3D model point for each 2D pixel of the object. These dense 2D-3D correspondences are either obtained using UV maps \cite{DPOD} or regressing the coordinates in the object's 3D model space \cite{Pix2Pose}\cite{CDPN}. The 6D poses are computed afterwards using PnP and RANSAC. DPOD\cite{DPOD} uses an additional refinement network that is fed with the cropped image patch of the object and another image patch that has to be rendered separately using the predicted pose from the first stage and outputs the refined pose.\\

While those works often report fast inference times for single object pose estimation, due to their indirect pose estimation approach using intermediate representations and computing the 6D pose subsequently for each object independently, the runtime is highly dependent of the number of objects per image. Furthermore, some methods can't handle multiple objects well and need a separate trained model for each object \cite{Pix2Pose}\cite{DOPE} or have problems with multiple instances in some cases and need additional modifications to handle these scenarios \cite{DPOD}. There are also some methods that rely on an external 2D object detector first to detect the objects of interest in the input image and to operate on these detections separately \cite{CDPN}\cite{Pix2Pose}. All these mentioned cases increase the complexity of the approaches and limit their applicability in some use cases, especially when multiple objects or instances are involved.

\subsection{2D Object Detection}
\label{subsection_2d_object_detection}
While the development from R-CNN\cite{RCNN} over Fast-R-CNN\cite{FastRCNN} to Faster-R-CNN\cite{FasterRCNN} led to substantial gains in accuracy and performance in the field of 2D object detection, those so-called two-stage approaches tend to be more complex and not as efficient as one-stage methods \cite{EfficientDet}. Nevertheless, they usually achieved a higher accuracy under similar computational costs when compared to one-stage methods \cite{RetinaNet}. The difference between both is that one-stage detectors perform the task in a single shot, while two-stage approaches perform a region proposal step in the first stage and make the final object detection in the second step based on the region proposals. Since RetinaNet\cite{RetinaNet} closed the accuracy gap, one-stage detectors gained more attention due to their simplicity and efficiency \cite{EfficientDet}. A common method to push the detection performance further, is to use larger backbone networks, like deeper ResNet\cite{ResNet} variants or AmoebaNet\cite{AmoebaNet}, or to increase the input resolution \cite{YOLOv3}\cite{AmoebaNet_NAS_FPN}. Yet, with the gains in detection accuracy, the computational costs often significantly increase in parallel, which reduces their applicability to use cases without computational constraints. Therefore, Tan \etal\cite{EfficientDet} focused not only on accuracy but also on efficiency and brought the idea of the scalable backbone architecture EfficientNet\cite{EfficientNet} to 2D object detection. The resulting EfficientDet architecture family can be scaled easily with a single hyperparameter over a wide range of computational resources - from mobile size to a huge network achieving state-of-the-art result on COCO test-dev\cite{COCO}. To introduce those advantages also to the field of 6D object pose estimation, we therefore base our approach on this architecture.

\section{Methods}
\label{section_methods}

\begin{figure*}
\centering
\includegraphics[width = 0.95\textwidth]{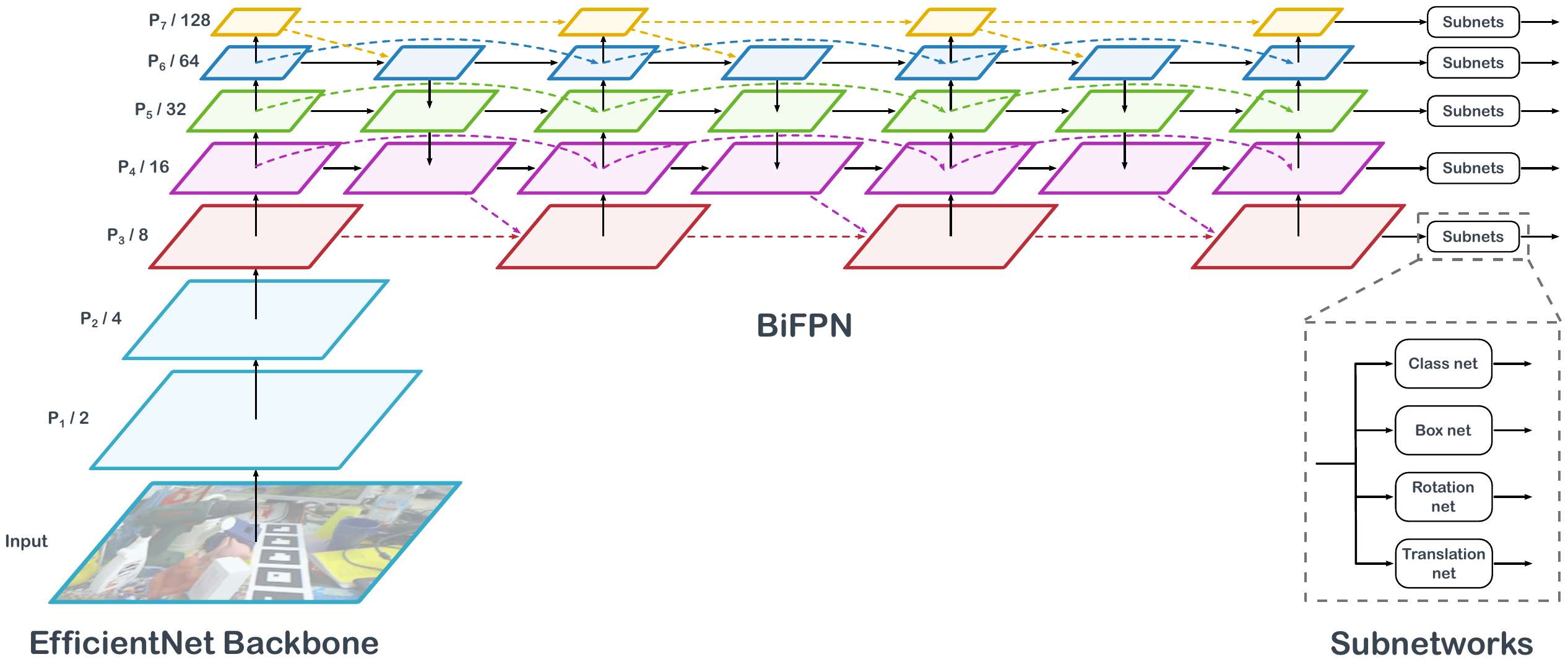}
\caption{Schematic representation of our EfficientPose architecture including the EfficientNet\cite{EfficientNet} backbone, the bidirectional feature pyramid network (BiFPN) and the prediction subnetworks.}
\label{figure_efficientpose_architecture}
\end{figure*}

In this section we describe our approach for 6D object pose estimation using RGB images as input. The complete 6D pose is composed of two parts - the 3D rotation $\mathbf{R} \in SO(3)$ of the object and the 3D translation $\textbf{t} \in \mathbb{R}^3$. This 6D pose represents the rigid transformation from the object coordinate system into the camera coordinate system. Because this overall task involves several subtasks like detecting objects in the 2D image first, handling multiple object categories and instances, etc. which are already solved in recent works from the relatively matured field of 2D object detection, we decided to base our work on such an 2D object detection approach and extend it with the ability to also predict the 6D pose of objects.

\subsection{Extending the EfficientDet architecture}
\label{subsection_extending_efficientdet}

Our goal is to extend the EfficientDet architecture in an intuitive way and keep the computational overhead rather small. Therefore, we add two new subnetworks, analogous to the classification and bounding box regression subnetworks, but instead of predicting the class and bounding box offset for each anchor box, the new subnets predict the rotation $\textbf{R}$ and translation $\textbf{t}$ respectively. Since those subnets are small and share the input feature maps with the already existing classification and box subnets, the additional computational cost is minimal. Integrating the task of 6D pose estimation via those two subnetworks and using the anchor box mapping and non-maximum-suppression (NMS) of the base architecture to filter out background and multiple detections, we are able to create an architecture that can detect the 
\begin{itemize}
	\item Class
	\item 2D bounding box
	\item Rotation
	\item Translation
\end{itemize}
of one or more object instances and categories for a given RGB image in a single shot. To maintain the scalability of the underlying EfficientDet architecture, the size of the rotation and translation network is also controlled by the scaling hyperparameter $\phi$. A high-level view of our architecture is presented in \autoref{figure_efficientpose_architecture}. For further information about the base architecture we refer the reader to the EfficientDet publication\cite{EfficientDet}.

\subsection{Rotation Network}
\label{subsection_rotation_network}

\begin{figure*}
\centering
\includegraphics[width = 0.95\textwidth]{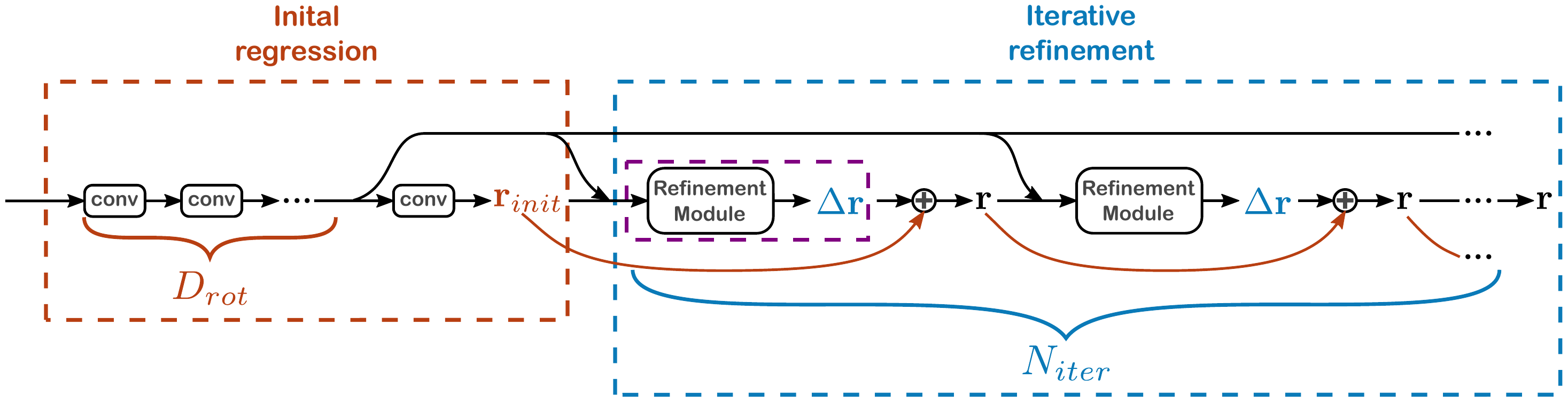}
\caption{Rotation network architecture with the initial regression and iterative refinement module. Each conv block consists of a depthwise separable convolution layer followed by group normalization and SiLU activation.}
\label{figure_rotation_architecture}
\end{figure*}

\begin{figure}
\centering
\includegraphics[width = 0.35\columnwidth]{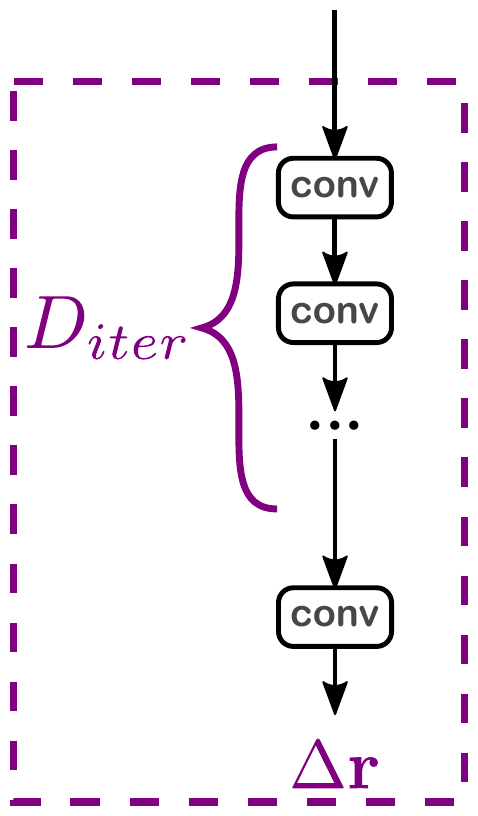}
\caption{Architecture of the rotation refinement module. Each conv block consists of a depthwise separable convolution layer followed by group normalization and SiLU activation.}
\label{figure_refinement_architecture}
\end{figure}

We choose axis angle representation for the rotation because it needs fewer parameters than quaternions and Mahendran \etal\cite{axis_angle_vs_quaternion} found that it also performed slightly better in their experiments. Yet, this representation is not crucial for our approach and can also be switched if needed. So instead of a rotation matrix $\mathbf{R} \in SO(3)$, the subnetwork predicts one rotation vector $\textbf{r}\in\mathbb{R}^{3}$ for each anchor box. The network architecture is similar to the classification and box network in EfficientDet\cite{EfficientDet} but instead of using the output $\textbf{r}_{init}$ directly as the regressed rotation, we further add an iterative refinement module, inspired by Kanazawa \etal\cite{HMR}. This module takes the concatenation along the channel dimension of the current rotation $\textbf{r}_{init}$ and the output of the last convolution layer prior to the initial regression layer which outputs $\textbf{r}_{init}$ as the input and regresses $\Delta \textbf{r}$ so that the final rotation regression is 
\begin{equation}
\label{equation_rotation_r}
\textbf{r} = \textbf{r}_{init} + \Delta \textbf{r}
\end{equation}
The iterative refinement module consists of $D_{iter}$ depthwise separable convolution layer\cite{DepthwiseSeparableConv}, each layer followed by group normalization \cite{GroupNorm} and SiLU (swish-1) activation function \cite{swish_1}\cite{swish_2}\cite{swish_3}. The number of layers $D_{iter}$, dependent by the scaling hyperparameter $\phi$ is described by the following equation
\begin{equation}
\label{equation_d_iter}
D_{iter}(\phi) = 2 + \lfloor \phi / 3 \rfloor
\end{equation}
where $\lfloor \rfloor$ denotes the floor function. These layers are followed by the output layer - a single depthwise separable convolution layer with linear activation function - which outputs $\Delta \textbf{r}$.\\

This iterative refinement module is applied $N_{iter}$ times to the rotation $\textbf{r}$, initialized with the output of the base network $\textbf{r}_{init}$ and after each intermediate iteration step $\textbf{r}$ is set to $\textbf{r}_{init}$ for the next step. $N_{iter}$ is also dependent on $\phi$ to preserve the scalability and is defined as follows
\begin{equation}
\label{equation_n_iter}
N_{iter}(\phi) = 1 + \lfloor \phi / 3 \rfloor
\end{equation}
The number of channels for all layers are the same as in the class and box networks, except for the output layers, which are determined by the number of anchors and rotation parameters. \autoref{equation_d_iter} and \autoref{equation_n_iter} are based on the equation for the depth $D_{box}$ and $D_{class}$ of the box and class networks from EfficientDet\cite{EfficientDet} but are not backed up with further experiments and could possibly be optimized. The architecture of the complete rotation network is presented in \autoref{figure_rotation_architecture}, while the detailed topology of the refinement module is shown in \autoref{figure_refinement_architecture}.\\

Even though our design of the rotation and translation network, described in \autoref{subsection_translation_network}, is based on the box and class network from the vanilla EfficientDet, we replace batch normalization with group normalization to reduce the minimum needed batch size during training \cite{GroupNorm}. With this replacement we are able to successfully train the rotation and translation network from scratch with a batch size of 1 which heavily reduces the needed amount of memory during training compared to the needed minimum batch size of 32 with batch normalization. We aim for 16 channels per group which works well according to Wu \etal\cite{GroupNorm} and therefore calculating the number of groups $N_{groups}$ as follows
\begin{equation}
\label{equation_num_groups}
N_{groups}(\phi) = \lfloor \frac{W_{bifpn}(\phi)}{16} \rfloor
\end{equation}
where $W_{bifpn}$ denotes the number of channels in the EfficientDet BiFPN and prediction networks \cite{EfficientDet}.

\subsection{Translation Network}
\label{subsection_translation_network}
The network topology of the translation network is basically the same as for the rotation network described in \autoref{subsection_rotation_network}, with the difference of outputting a translation $\textbf{t} \in \mathbb{R}^{3}$ for each anchor box. However, instead of directly regressing all components of the translation vector $\textbf{t} = (t_x, t_y, t_z)^T$, we adopt the approach of PoseCNN\cite{PoseCNN} and split the task into predicting the 2D center point $\textbf{c} = (c_x, c_y)^T$ of the object in pixel coordinates and the distance $t_z$ separately. With the center point $\textbf{c}$, the distance $t_z$ and the intrinsic camera parameters, the missing components $t_x$ and $t_y$ of the translation $\textbf{t}$ can be calculated using the following equations assuming a pinhole camera
\begin{equation}
\label{equation_tx}
t_x = \frac{(c_x - p_x) \cdot t_z}{f_x}
\end{equation}

\begin{equation}
\label{equation_ty}
t_y = \frac{(c_y - p_y) \cdot t_z}{f_y}
\end{equation}

where $\textbf{p} = (p_x, p_y)^T$ is the principal point and $f_x$ and $f_y$ are the focal lengths. \\

\begin{figure}
\centering
\includegraphics[width = 0.9\columnwidth]{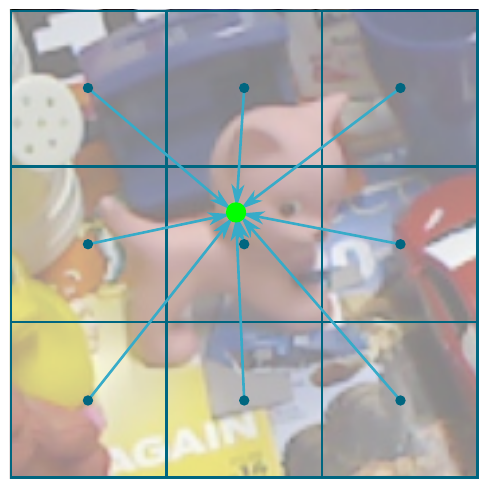}
\caption{Illustration of the 2D center point estimation process. The target for each point in the feature map is the offset from the current location to the object's center point.}
\label{figure_translation_centerpoint}
\end{figure}

For each anchor box we predict the offset in pixels from the center of this anchor box to the center point $\textbf{c}$ of the corresponding object. This is equivalent to predicting the offset to the center point from the current point in the given feature map, as illustrated in \autoref{figure_translation_centerpoint}. To maintain the relative spatial relations, the offset is normalized with the stride of the input feature map from every level of the feature pyramid. Using the 
\begin{itemize}
	\item predicted relative offsets,
	\item the coordinate maps $\textbf{X}$ and $\textbf{Y}$ of the feature maps where every point contains its own $x$ and $y$ coordinate respectively
	\item and the strides,
\end{itemize}
the absolute coordinates of the center point $\textbf{c}$ can be calculated. Our intention here is that it might be easier for the network to predict the relative offset at each point in the feature maps instead of directly regressing the absolute coordinates $c_x$ and $c_y$ due to the translational invariance of the convolution. We also verified this assumption experimentally.\\

The above described calculations of the translation $\textbf{t}$ from the 2D center point $\textbf{c}$ and the depth $t_z$, as well as the absolute center point coordinates $c_x$ and $c_y$ from their predicted relative offsets are both implemented in separate TensorFlow\cite{TensorFlow} layers to avoid extra post-processing steps and to enable GPU or TPU acceleration, while keeping the architecture as simple as possible. As mentioned earlier, the calculation of $\textbf{t}$ also needs the intrinsic camera parameters which is the reason why there is another input layer needed for the translation network. This input layer provides a vector $\textbf{a} \in \mathbb{R}^6$ for each input image which contains the focal lengths $f_x$ and $f_y$ of the pinhole camera, the principal point coordinates $p_x$ and $p_y$ and finally an optional translation scaling factor $s_{translation}$ and the image scale $s_{image}$. The translation scaling factor $s_{translation}$ can be used to adjust the translation unit, e.g. from mm to m. The image scale $s_{image}$ is the scaling factor from the original image size to the input image size which is needed to rescale the predicted center point $\textbf{c}$ to the original image resolution to apply \autoref{equation_tx} and \autoref{equation_ty} for recovering $\textbf{t}$.

\subsection{Transformation Loss}
\label{subsection_transformation_loss}
The loss function we use is based on the PoseLoss and ShapeMatch-Loss from PoseCNN\cite{PoseCNN} but instead of considering only the rotation, our approach takes also the translation into account. For asymmetric objects our loss $L_{asym}$ is defined as follows
\begin{equation}
\label{equation_asym_loss}
\begin{split}
L_{asym} = \frac{1}{m}\sum_{\mathbf{x} \in \mathcal{M}}\| (\text{Rot}(\mathbf{\tilde{r}}, \mathbf{x}) + \mathbf{\tilde{t}}) \\
-  (\text{Rot}(\mathbf{r}, \mathbf{x}) + \mathbf{t})  \|_2,
\end{split}
\end{equation}
whereby $\text{Rot}(\mathbf{r}, \mathbf{x})$ and $\text{Rot}(\mathbf{\tilde{r}}, \mathbf{x})$ respectively indicate the rotation of $\mathbf{x}$ with the ground truth rotation $\mathbf{r}$ and the estimated rotation $\mathbf{\tilde{r}}$ by applying the Rodrigues' rotation formula \cite{RodriguesRotation_1}\cite{RodriguesRotation_2}. Furthermore, $\mathcal{M}$ denotes the set of the object's 3D model points and $m$ is the number of points. The loss function basically performs the transformation of the object of interest with the ground truth 6D pose and the estimated 6D pose and then calculates the mean point distances between the transformed model points which is identical to the ADD metric described in \autoref{subsection_evaluation_metrics}. This approach has the advantage that the model is directly optimized on the metric with which the performance is measured. It also eliminates the need of an extra hyperparameter to balance the partial losses when the rotation and translation losses are calculated independently from each other.\\

To also handle symmetric objects, the corresponding loss $L_{sym}$ is given by the following equation
\begin{equation}
\label{equation_sym_loss}
\begin{split}
L_{sym} = \frac{1}{m}\sum_{\mathbf{x}_1 \in \mathcal{M}} \min_{\mathbf{x}_2 \in \mathcal{M}} \| (\text{Rot}(\mathbf{\tilde{r}}, \mathbf{x}_1) + \mathbf{\tilde{t}}) \\
 - (\text{Rot}(\mathbf{r}, \mathbf{x}_2) + \mathbf{t})  \|_2
\end{split}
\end{equation}
which is similar to $L_{asym}$ but instead of strictly calculating the distance between the matching points of the two transformed point sets, the minimal distance for each point to any point in the other transformed point set is taken into account. This helps to avoid unnecessary penalization during training when dealing with symmetric objects as described by Xiang \etal\cite{PoseCNN}.\\

The complete transformation loss function $L_{trans}$ is defined as follows
\begin{equation}
\label{equation_transformation_loss}
L_{trans} = 
\begin{cases}
L_{sym} & \text{if symmetric}, \\
L_{asym} & \text{if asymmetric}.
\end{cases}
\end{equation}

\subsection{6D Augmentation}
\label{subsection_6d_augmentation}

\begin{figure}[b]
\includegraphics[width = \linewidth]{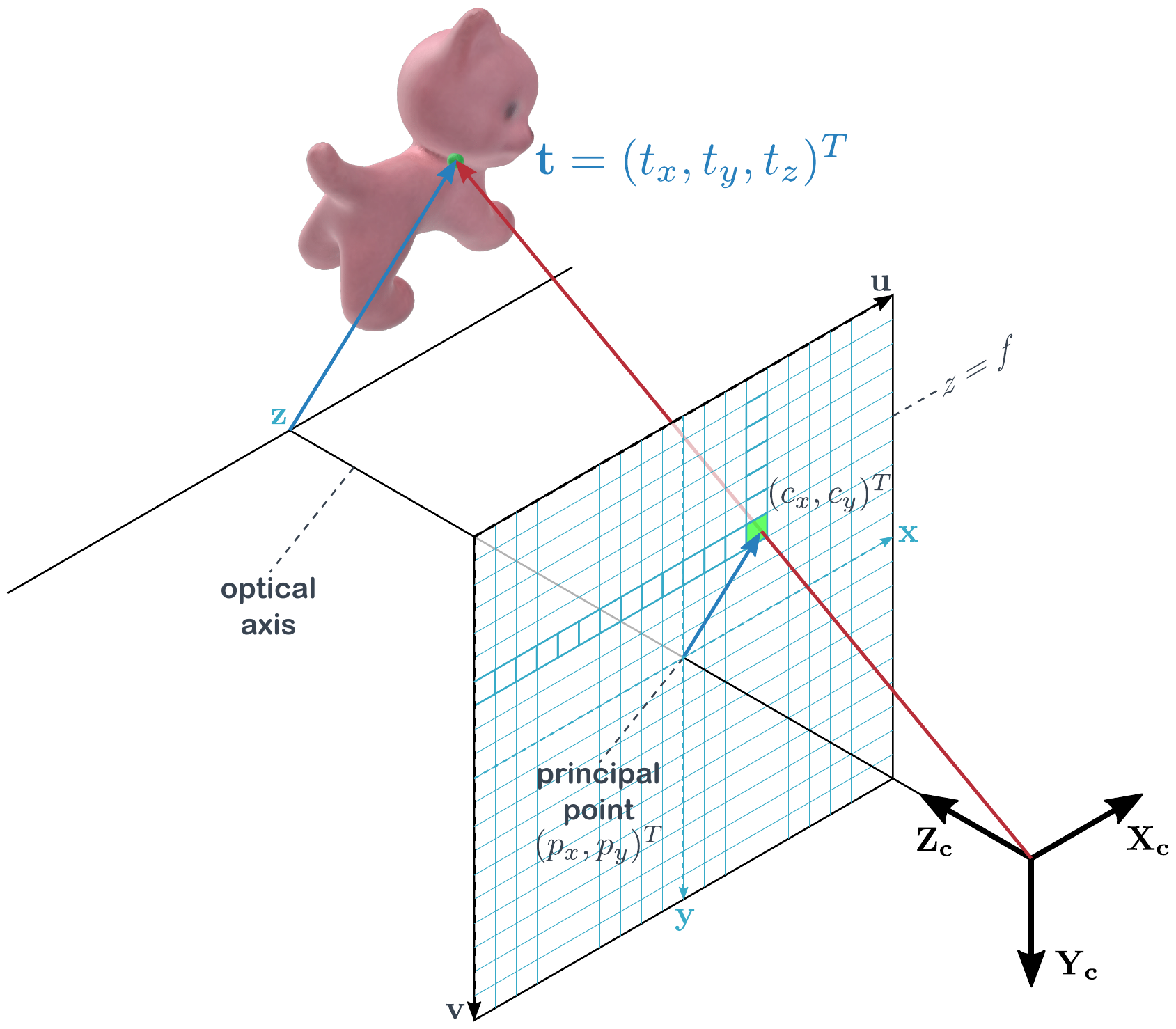}
\caption{Schematic figure of a pinhole camera illustrating the projection of an object's 3D center point onto the 2D image plane.}
\label{figure_camera_3d_2d_projection}
\end{figure}

The Linemod\cite{Linemod} and Occlusion\cite{Occlusion} datasets used in this work are very limited in the amount of annotated data. Linemod roughly consists of about 1200 annotated examples per object and Occlusion is a subset of Linemod where all objects of a single scene are annotated so the amount of data is equally small. This makes it especially hard for large neural networks to converge to more general solutions. Data augmentation can help a lot in such scenarios\cite{RandAugment}\cite{AmoebaNet_NAS_FPN} and methods which rely on any 2D detection and PnP approach have a great advantage here. Such methods can easily use image manipulation techniques like rotation, scaling, shearing etc. because the 2D targets, e.g. keypoints, can be relatively easy transformed according to the image transformation. Approaches that directly predict the 6D pose of an object are limited in this aspect because some image transformations, like rotation for example, lead to a mismatch between image and ground truth 6D pose. To overcome this issue, we developed a 6D augmentation that is able to rotate and scale an image randomly and transform the ground truth 6D poses so they still match to the augmented image.\\
As can be seen in \autoref{figure_camera_3d_2d_projection}, when performing a 2D rotation of the image around the principal point with an angle $\theta \in [0^{\circ}, 360^{\circ})$, the 3D rotation $\mathbf{R}$ and translation $\mathbf{t}$ of the 6D pose also have to be rotated with $\theta$ around the $z$-axis. This rotation around the $z$-axis can be described with the rotation vector $\Delta \mathbf{r}$ in axis angle representation as follows
\begin{equation}
\label{equation_aug_delta_r}
\Delta \mathbf{r} = (0, 0, \frac{\theta}{180 \cdot \pi})^T.
\end{equation}
Using the rotation matrix $\Delta \mathbf{R}$ obtained from $\Delta \mathbf{r}$, the augmented rotation matrix $\mathbf{R}_{aug}$ and translation $\mathbf{t}_{aug}$ can be computed with the following equations
\begin{equation}
\label{equation_r_aug}
\mathbf{R}_{aug} = \Delta \mathbf{R} \cdot \mathbf{R}
\end{equation}

\begin{equation}
\label{equation_t_aug}
\mathbf{t}_{aug} = \Delta \mathbf{R} \cdot \mathbf{t}
\end{equation}

To handle image scaling as an additional augmentation technique as well, we need to adjust the $t_z$ component of the translation $\mathbf{t} = (t_x, t_y, t_z)^T$. Rescaling the image with a factor $f_{scale}$, the augmented translation $\mathbf{t}_{aug}$ can be calculated as follows
\begin{equation}
\label{equation_t_aug_scale}
\mathbf{t}_{aug} = (t_x, t_y, \frac{t_z}{f_{scale}})^T.
\end{equation}

\begin{figure}
\includegraphics[width = \linewidth]{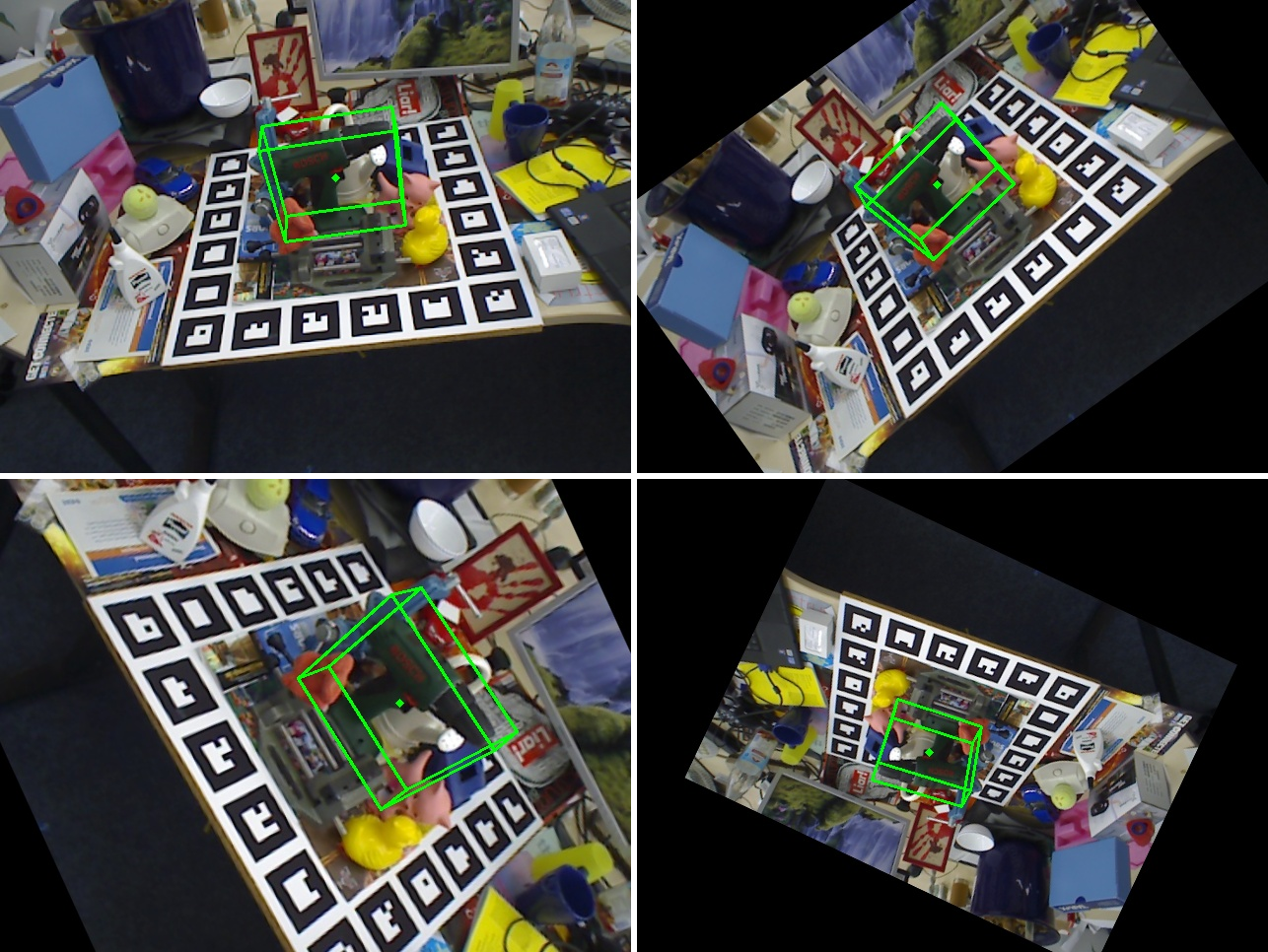}
\caption{Some examples of our proposed 6D augmentation. The image in the top left is the original image with the projected object cuboid, transformed with the ground truth 6D pose. The other images are obtained through augmenting the image and the 6D poses separately from each other and then transforming the object's cuboid with the augmented 6D poses and finally project each cuboid onto the corresponding augmented image.}
\label{figure_6d_aug_examples}
\end{figure}

It has to be mentioned that the scaling augmentation introduces an error if the object of interest is not in the image center. When rescaling the image, the 2D projection of the object remains the same. It only becomes bigger or smaller. However, when moving the object along the $z$-axis in reality, the view from the camera to the 3D object would change and so the projection onto the 2D image plane. Nevertheless, the benefits from the additional data obtained with this augmentation technique strongly outweigh its introduced error as shown in \autoref{subsection_ablation_study}. \autoref{figure_6d_aug_examples} contains some examples where the top left image is the original image with the ground truth 6D pose and the other images are augmented with the method described in this subsection. In this work we use for all experiments a random angle $\theta$, uniformly sampled from the interval $[0^{\circ}, 360^{\circ})$ and a random scaling factor $f_{scale}$, uniformly sampled from $[0.7, 1.3]$.

\subsection{Color space Augmentation}
\label{subsection_colorspace_augmentation}
We also use several augmentation techniques in the color space that can be applied without further need to adjust the annotated 6D poses. For this task we adopt the RandAugment\cite{RandAugment} method which is a learned augmentation that is able to boost performance and enhance generalization among several datasets and models. It consists of multiple augmentation methods, like adjusting the contrast and brightness of the input image, and can be tuned with two parameters - the number $n$ of applied image transformations and the strength $m$ of these transformations.\\
As mentioned earlier, some image transformations like rotation and shearing lead to a mismatch between the input image and the ground truth 6D poses, so we remove those augmentation techniques from the RandAugment method. We further add gaussian noise to the selection. To maintain the approach of setting the augmentation strength with the parameter $m$, the channel-wise additive gaussian noise is sampled from a normal distribution with the range $[0, \frac{m}{100} \cdot 255]$.\\
For all our experiments we choose $n$ randomly sampled from an integer uniform distribution $[1, 3]$ and $m$ from $[1, 14]$ for each image.

\section{Experiments}
\label{section_experiments}
In this section we describe the experiments we did, our experimental setup with implementation details as well as the evaluation metrics we use. In case of the Linemod experiment, we also compare our results to current state-of-the-art methods. Please note that our approach can be scaled from $\phi = 0$ to $\phi = 7$ in integer steps but due to computational constraints, we only use $\phi = 0$ and $\phi = 3$ in our experiments.

\subsection{Datasets}
\label{subsection_datasets}
We evaluate our approach on two popular benchmark datasets which are described in this subsection.

\subsubsection{Linemod}
\label{subsubsection_linemod}
The Linemod\cite{Linemod} dataset is a popular and widely-used benchmark dataset for 6D object pose estimation. It consists of 13 different objects (actually 15 but only 13 are used in most other works \cite{YOLO6D}\cite{Pix2Pose}\cite{PVNet}\cite{DPOD}\cite{CDPN}\cite{HybridPose}) which are placed in 13 cluttered scenes. For each scene only one object is annotated with it's 6D pose although other objects are visible at the same time. So despite of our approach being able to detect multiple objects and to estimate their poses, we had to train one model for each object. There are about 1200 annotated examples per object and we use the same train and test split as other works \cite{Linemod_train_test_split}\cite{PVNet}\cite{YOLO6D} for fair comparison. This split selects training images so the object poses had a minimum angular distance of 15$^\circ$, which results in about 15\% training images and 85\% test images. Furthermore, we do not use any synthetically rendered images for training. We compare our results with state-of-the-art methods in \autoref{subsection_comparison_linemod}.

\begin{table*}
\begin{tabularx}{\textwidth}{ | r | X  X  X  X  X  X  X | X  X |}
 \hline
 Method & YOLO6D \cite{YOLO6D} & Pix2Pose \cite{Pix2Pose} & PVNet \cite{PVNet} & DPOD \cite{DPOD} & DPOD+ \cite{DPOD} & CDPN \cite{CDPN} & Hybrid- Pose \cite{HybridPose} & \textbf{Ours} $\phi = 0$ & \textbf{Ours} $\phi = 3$  \\
 \hline
 ape            & 21.62 & 58.1 & 43.62 & 53.28 & 87.73 & 64.38 & 63.1 & 87.71 & \textbf{89.43} \\
 benchvise      & 81.80 & 91.0 & \textbf{99.90} & 95.34 & 98.45 & 97.77 & \textbf{99.9} & 99.71 & 99.71 \\
 cam            & 36.57 & 60.9 & 86.86 & 90.36 & 96.07 & 91.67 & 90.4 & 97.94 & \textbf{98.53} \\
 can            & 68.80 & 84.4 & 95.47 & 94.10 & \textbf{99.71} & 95.87 & 98.5 & 98.52 & 99.70 \\
 cat            & 41.82 & 65.0 & 79.34 & 60.38 & 94.71 & 83.83 & 89.4 & \textbf{98.00} & 96.21 \\
 driller        & 63.51 & 76.3 & 96.43 & 97.72 & 98.80 & 96.23 & 98.5 & \textbf{99.90} & 99.50 \\
 duck           & 27.23 & 43.8 & 52.58 & 66.01 & 86.29 & 66.76 & 65.0 & \textbf{90.99} & 89.20 \\
 eggbox*        & 69.58 & 96.8 & 99.15 & 99.72 & 99.91 & 99.72 & \textbf{100} & \textbf{100} & \textbf{100} \\
 glue*          & 80.02 & 79.4 & 95.66 & 93.83 & 96.82 & 99.61 & 98.8 & \textbf{100} & \textbf{100} \\
 holepuncher    & 42.63 & 74.8 & 81.92 & 65.83 & 86.87 & 85.82 & 89.7 & 95.15 & \textbf{95.72} \\
 iron           & 74.97 & 83.4 & 98.88 & 99.80 & \textbf{100} & 97.85 & \textbf{100} & 99.69 & 99.08 \\
 lamp           & 71.11 & 82.0 & 99.33 & 88.11 & 96.84 & 97.89 & 99.5 & \textbf{100} & \textbf{100} \\
 phone          & 47.74 & 45.0 & 92.41 & 74.24 & 94.69 & 90.75 & 94.9 & 97.98 & \textbf{98.46} \\
 \hline
 Average & 55.95 & 72.4 & 86.27 & 82.98 & 95.15 & 89.86 & 91.3 & \textbf{97.35} & \textbf{97.35} \\
 \hline
\end{tabularx}
\caption{Quantitative evaluation and comparison on the Linemod dataset in terms of the ADD(-S) metric. Symmetric objects are marked with * and approaches marked with + are using an additional refinement method.}
\label{table_linemod_comparison}
\end{table*}

\subsubsection{Occlusion}
\label{subsubsection_occlusion}
The Occlusion dataset is a subset of Linemod and consists of a single scene of Linemod where eight other objects visible in this scene are additionally annotated. These objects are partially heavily occluded which makes it challenging to estimate their 6D poses. We use this dataset to evaluate our method's ability for multi object 6D pose estimation. Therefore, we trained a single model on the Occlusion dataset. We use the same train and test split as for the corresponding Linemod scene. The results of this experiment are presented in \autoref{subsection_occlusion_single_model}.\\
Please note that the evaluation convention in other works \cite{PoseCNN}\cite{PVNet} is to use the Linemod dataset for training and the complete Occlusion data as the test set, so this experiment is not comparable with those works.

\subsection{Evaluation metrics}
\label{subsection_evaluation_metrics}
We evaluate our approach with the commonly used ADD(-S) metric\cite{ADD}. This metric calculates the average point distances between the 3D model point set $\mathcal{M}$ transformed with the ground truth rotation $\mathbf{R}$ and translation $\mathbf{t}$ and the model point set transformed with the estimated rotation $\mathbf{\tilde{R}}$ and translation $\mathbf{\tilde{t}}$. It also differs between asymmetric and symmetric objects. For asymmetric objects the ADD metric is defined as follows
\begin{equation}
\label{equation_add_metric}
\text{ADD} = \frac{1}{m}\sum_{\mathbf{x} \in \mathcal{M}}\| (\mathbf{Rx} + \mathbf{t}) - (\mathbf{\tilde{R}x} + \mathbf{\tilde{t}})  \|_2.
\end{equation}
An estimated 6D pose is considered correct if the average point distance is smaller than 10\% of the object's diameter.\\
Symmetric objects are evaluated using the ADD-S metric which is given by the following equation
\begin{equation}
\label{equation_add_s_metric}
\begin{split}
\text{ADD-S} = \frac{1}{m}\sum_{\mathbf{x}_1 \in \mathcal{M}} \min_{\mathbf{x}_2 \in \mathcal{M}} \| (\mathbf{Rx} + \mathbf{t})\\
 - (\mathbf{\tilde{R}x} + \mathbf{\tilde{t}})  \|_2.
\end{split}
\end{equation}
Finally, the ADD(-S) metric is defined as 
\begin{equation}
\label{equation_complete_add_s_metric}
\text{ADD(-S)} = 
\begin{cases}
\text{ADD} & \text{if asymmetric}, \\
\text{ADD-S} & \text{if symmetric}.
\end{cases}
\end{equation}

\subsection{Implementation Details}
\label{subsection_implementation_details}
We use the Adam optimizer\cite{Adam} with an initial learning rate of 1e-4 for all our experiments and a batch size of 1. We also use gradient norm clipping with a threshold of 0.001 to increase training stability. The learning rate is reduced with a factor of 0.5 if the average point distance does not decrease within the last 25 evaluations on the test set. The minimum learning rate is set to 1e-7. Since the training set of Linemod and Occlusion is very small (roughly 180 examples per object), as mentioned in \autoref{subsubsection_linemod} and \autoref{subsubsection_occlusion}, we evaluate our model only every 10 epochs to measure training progression. Our model is trained for 5000 epochs. The complete loss function $L$ is composed of three parts - the classification loss $L_{class}$, the bounding box regression loss $L_{bbox}$ and the transformation loss $L_{trans}$. To balance the influence of these partial losses on the training procedure, we introduce a hyperparameter $\lambda$ for each partial loss, so the final loss $L$ is defined as follows
\begin{equation}
\label{equation_complete_loss}
L = \lambda_{class} \cdot L_{class} + \lambda_{bbox} \cdot L_{bbox} + \lambda_{trans} \cdot L_{trans}
\end{equation}
We found that $\lambda_{class} = \lambda_{bbox} = 1$ and $\lambda_{trans} = 0.02$ performs well in our experiments. To calculate the transformation loss $L_{trans}$, described in \autoref{subsection_transformation_loss}, we use $m = 500$ points of the 3D object model point set $\mathcal{M}$.\\
We use our 6D and color space augmentation by default with the parameters mentioned in \autoref{subsection_6d_augmentation} and \autoref{subsection_colorspace_augmentation} respectively but randomly skip augmentation with a probability of 0.02 to also include examples from the original image domain in our training process.\\
We initialize the neural network, except the rotation and translation network, with COCO\cite{COCO} pretrained weights from the vanilla EfficientDet\cite{EfficientDet}. Because of our small batch size, we freeze all batch norm layers during training and use the population statistics learned from COCO.

\subsection{Comparison on Linemod}
\label{subsection_comparison_linemod}

\begin{figure}[h]
\includegraphics[width = \linewidth]{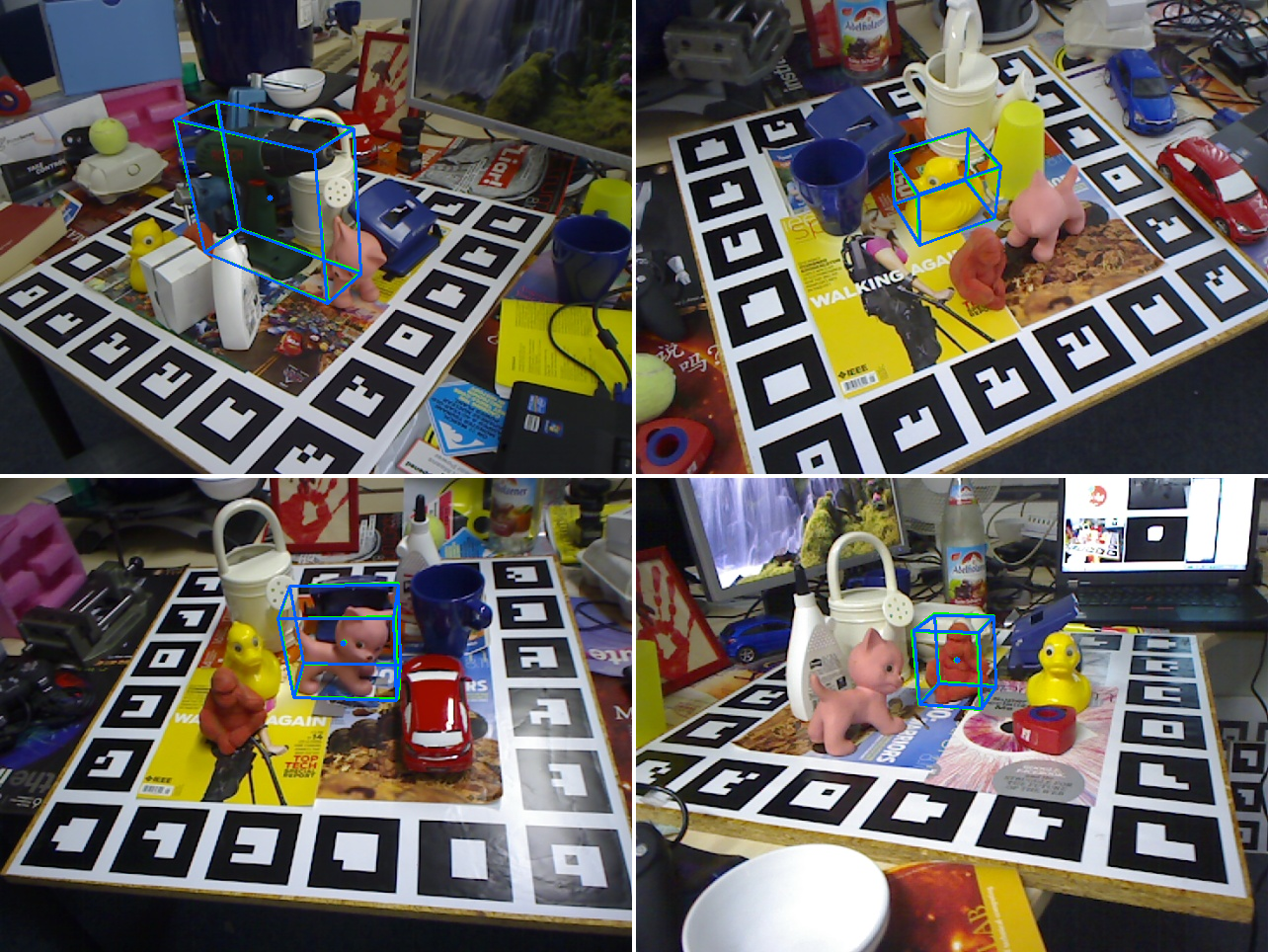}
\caption{Some example predictions for qualitative evaluation of our $\phi = 0$ model on the Linemod test dataset. Green 3D bounding boxes visualize ground truth poses while our estimated poses are represented by blue boxes.}
\label{figure_merged_linemod_phi_0_example}
\end{figure}

In \autoref{table_linemod_comparison} we compare our results with current state-of-the-art methods using RGB input on the Linemod dataset in terms of the ADD(-S) metric. Our approach outperforms all other methods without further refinement steps by a large margin. Even DPOD+ which uses an additional refinement network and reported the best results on Linemod so far using only RGB input data, is outperformed considerably by our method, roughly halving the remaining error. Note again that, in contrast to all other recent works in \autoref{table_linemod_comparison} \cite{YOLO6D}\cite{Pix2Pose}\cite{PVNet}\cite{DPOD}\cite{CDPN}\cite{HybridPose}, our approach detects and estimates objects with their 6D poses in a single shot without the need of further post-processing steps like RANSAC-based voting or PnP. This fact demonstrates the current domination of 2D+PnP approaches in the high accuracy regime on Linemod using only RGB input. Since a crucial part of our reported performance on Linemod is our proposed 6D augmentation, as can be seen in \autoref{subsection_ablation_study}, the question arises if the previous superiority of 2D+PnP approaches over direct 6D pose estimation comes from the broader use of some augmentation techniques like rotation, which better enriches the small Linemod dataset. To the best of our knowledge, our approach is the first holistic method achieving competitive performance on Linemod with current state-of-the-art approaches like PVNet\cite{PVNet}, DPOD\cite{DPOD} and HybridPose\cite{HybridPose}. We therefore demonstrate that single shot direct 6D object pose estimation approaches are able to compete in terms of accuracy with 2D+PnP approaches and even with additional refinement methods. \autoref{figure_merged_linemod_phi_0_example} shows some qualitative results of our method.
\\
Interestingly, the performance of our $\phi = 0$ and $\phi = 3$ models are nearly the same, despite of their different capacities. This suggests that the capacity of our $\phi = 0$ model is already enough for the single object 6D pose estimation task on Linemod and that the bottleneck seems to be the small amount of data. Additionally, the small $\phi = 0$ model may not suffer from overfitting as much as the larger models which could be an explanation why the $\phi = 0$ model performs slightly better on some objects. The advantage of the larger $\phi = 3$ model is much more pronounced at multi object 6D pose estimation as we demonstrate in \autoref{subsection_occlusion_single_model}.

\subsection{Multi object pose estimation}
\label{subsection_occlusion_single_model}

\begin{figure}
\includegraphics[width = \linewidth]{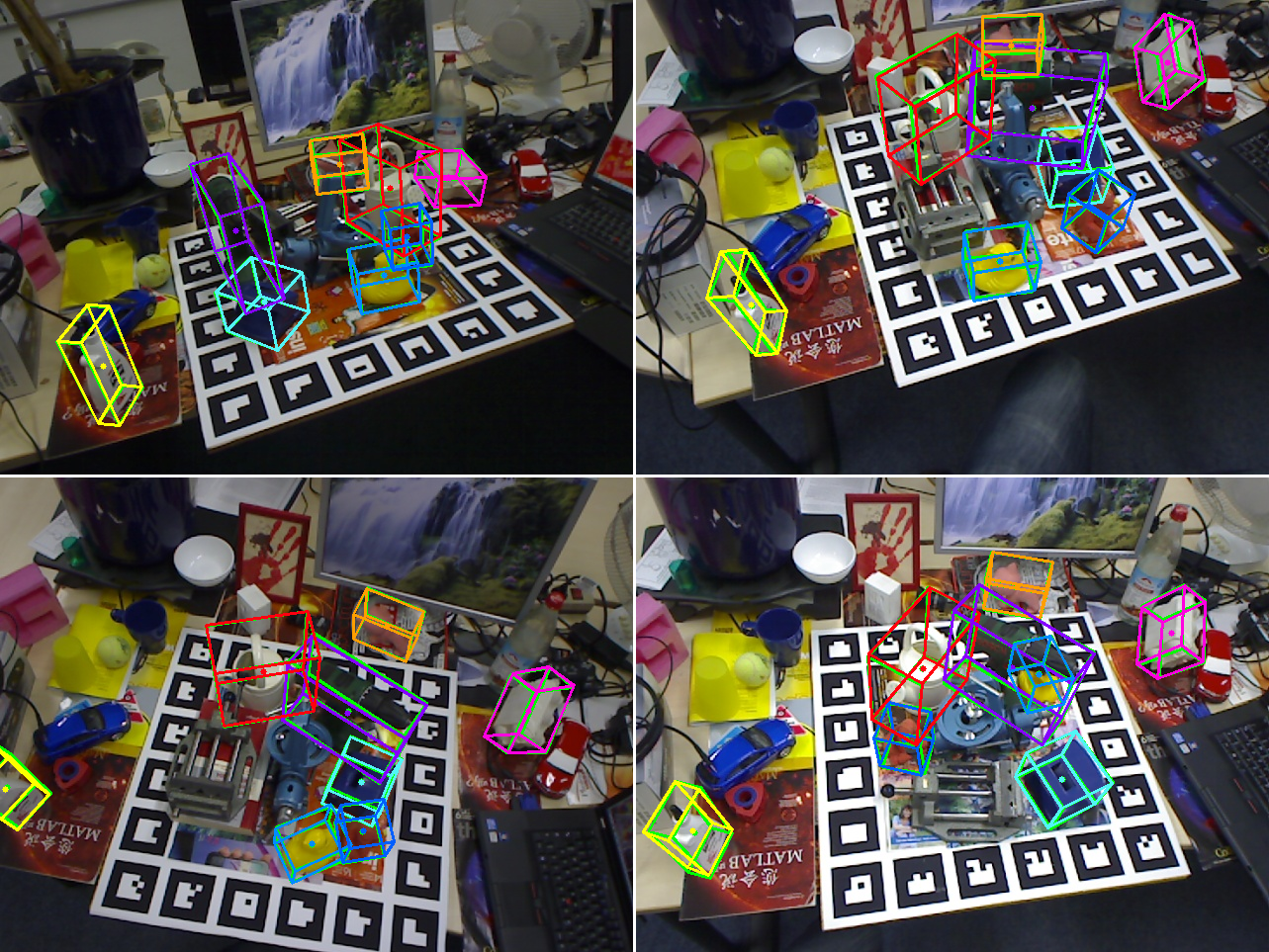}
\caption{Qualitative evaluation of our single $\phi = 0$ model's ability for estimating 6D poses of multiple objects in a single shot. Green 3D bounding boxes visualize ground truth poses while our estimated poses are represented by the other colors.}
\label{figure_merged_occlusion_phi_0_example}
\end{figure}

To validate that our approach is really capable of handling multiple objects in practice, we also trained a single model on Occlusion. Because of the reasons explained in \autoref{subsubsection_linemod}, we could not use the Linemod data of the objects for training like other works did \cite{PVNet}\cite{PoseCNN} and had to train our model on the Occlusion dataset. Therefore, we used the train and test split of the corresponding Linemod scene. Thus due to the different train and test data of this experiment, the reported results are not comparable to the results of other works \cite{PVNet}\cite{PoseCNN}. Training parameters remain the same as described in \autoref{subsection_implementation_details}.
\begin{table}
\begin{tabularx}{\columnwidth}{ | r | X  X |}
 \hline
 Method & \textbf{Ours} $\phi = 0$ & \textbf{Ours} $\phi = 3$  \\
 \hline
 ape            & 56.57 & \textbf{59.39} \\
 can            & 91.12 & \textbf{93.27} \\
 cat            & 68.58 & \textbf{79.78} \\
 driller        & 95.64 & \textbf{97.77} \\
 duck           & 65.31 & \textbf{72.71} \\
 eggbox*        & 93.46 & \textbf{96.18} \\
 glue*          & 85.15 & \textbf{90.80} \\
 holepuncher    & 76.53 & \textbf{81.95} \\
 \hline
 Average & 79.04 & \textbf{83.98} \\
 \hline
\end{tabularx}
\caption{Quantitative evaluation in terms of the ADD(-S) metric for the task of multi object 6D pose estimation using a single model on the Occlusion dataset. Symmetric objects are marked with *}
\label{table_occlusion_single_model_results}
\end{table}
 The results in \autoref{table_occlusion_single_model_results} suggest that our method is indeed able to detect and estimate the 6D poses of multiple objects in a single shot. \autoref{figure_title} and \autoref{figure_merged_occlusion_phi_0_example} are showing some examples with ground truth and estimated 6D poses of the Occlusion test set for qualitative evaluation. Interestingly the performance difference in terms of the ADD(-S) metric between the $\phi = 0$ and $\phi = 3$ model is quiet significant, unlike the Linemod experiment in \autoref{subsection_comparison_linemod}. We argue that the larger number of objects benefits more from the higher capacity of the $\phi = 3$ model. On top of that, the objects in this dataset often deal with severe occlusions which makes the 6D pose estimation task at the same time more challenging than on Linemod.

\subsection{Runtime analysis}
\label{subsection_runtime_analysis}

\begin{table*}
\centering
\begin{tabular}{ | c  r | c  c | c  c | c  c | c  c |}
 \hline
 \multicolumn{2}{|c|}{Method} & \multicolumn{4}{c|}{\textbf{Ours}} & \multicolumn{4}{c|}{{Vanilla EfficientDet\cite{EfficientDet}}}\\
 \hline
 \multicolumn{2}{|c|}{Model} & \multicolumn{2}{c|}{$\phi = 0$} & \multicolumn{2}{c|}{{$\phi = 3$}} & \multicolumn{2}{c|}{$\phi = 0$} & \multicolumn{2}{c|}{$\phi = 3$} \\
 \hline
	\multicolumn{2}{|c|}{Single or multiple objects}  & Single & Multi & Single & Multi & Single & Multi & Single & Multi \\
 \hline
 \multirow{2}{*}{Preprocessing} & ms & 8.17 & 8.12 & 24.38 & 24.26 & 8.07 & 8.56 & 25.69 & 26.95 \\
																& FPS & 122.40 & 123.14 & 41.02 & 41.22 & 123.92 & 116.82 & 38.93 & 37.11 \\
 \hline
 \multirow{2}{*}{Network} & ms & 28.18 & 29.96 & 81.60 & 82.69 & 19.26 & 21.42 & 51.71 & 53.97 \\
													& FPS & 35.49 & 33.38 & 12.26 & 12.09 & 51.91 & 46.69 & 19.34 & 18.53 \\
 \hline
 \multirow{2}{*}{End-to-end} 	& ms & 36.43 & 38.13 & 106.04 & 107.01 & 27.38 & 30.02 & 77.45 & 80.98 \\
															& FPS & 27.45 & 26.22 & 9.43 & 9.34 & 36.52 & 33.31 & 12.91 & 12.35 \\
 \hline
\end{tabular}
\caption{Runtime analysis and comparison of our method performing single and multiple object pose estimation while using different scales. For single object 6D pose estimation the Linemod dataset is used while for multi object pose estimation the Occlusion dataset is used which contains usually eight annotated objects per image. We further compare our method's runtime with the vanilla EfficientDet\cite{EfficientDet} to measure the influence of our 6D pose estimation extension.}
\label{table_runtime_analysis}
\end{table*}

In this subsection we examine the average runtime of our apprach in several scenarios and compare it with the vanilla EfficientDet\cite{EfficientDet}. The experiments were performed using the $\phi = 0$ and $\phi = 3$ model to study the influence of the scaling hyperparameter $\phi$. For each model we measured the runtime for single and multi object 6D pose estimation. To examine the single object task, we use the Linemod test dataset and for the latter the Occlusion test dataset because it typically contains eight annotated objects per image. All experiments were performed using a batch size of 1. We measured the time needed to
\begin{itemize}
\item preprocess the input data (Preprocessing),
\item the pure network inference time (Network)
\item and finally the complete end-to-end time including the data preprocessing, network inference with non-maximum-suppression and post-processing steps like rescaling the 2D bounding boxes to the original image resolution (end-to-end).
\end{itemize}
 To make a fair comparison with the vanilla EfficientDet, we use the same implementation on which our EfficientPose implementation is based on and also use the same weights so that the 2D detection remains identical. The results of these experiments are reported in \autoref{table_runtime_analysis}.\\
For a more fine grained evaluation, we performed a separate experiment in which we measured the runtime \wrt the number of objects per image. We used the Occlusion test set and cut out objects using the ground truth segmentation mask if necessary to match the target number of objects per image. Using this method we then iteratively measured the end-to-end runtime of the complete occlusion test set from a single object up to eight objects. To ensure a correct measuring, we filtered out images in which our model did not detect the estimated number of objects. The results of this experiment are visualized in \autoref{figure_title}. All experiments are run on the same machine with an i7-6700K CPU and a 2080 Ti GPU using Tensorflow 1.15.0, CUDA 10.0 and CuDNN 7.6.5.\\

Our $\phi = 0$ model runs end-to-end with an average 27.45 FPS at the single object 6D pose estimation task which makes it suitable for real time applications. Even more promising is the average end-to-end runtime of 26.22 FPS when performing multi object 6D pose estimation on the Occlusion test dataset which typically contains eight objects per image.\\ Using the much larger $\phi = 3$ model, our method still runs end-to-end at over 9 FPS while the difference between single and multi object 6D pose estimation nearly vanishes with 9.43 vs. 9.34 FPS. \autoref{figure_title} also demonstrates that the runtime of our approach is nearly independent from the number of objects per image. These results show the advantage of our method in multi object 6D pose estimation compared to the 2D detection approaches solving a PnP problem to obtain the 6D poses afterwards, which linearly increases the runtime with the number of objects. This makes our single shot approach very attractive for many real world scenarios, no matter if there are one or more objects.\\
When comparing the runtimes of the vanilla EfficientDet and our approach with roughly 35 vs. 27 FPS using $\phi = 0$ and 12 vs. 9 FPS with the $\phi = 3$ model, our extension of the EfficientDet architecture as described in \autoref{subsection_extending_efficientdet} seems computationally very efficient considering this rather small drop in frame rate in exchange for the additional ability of full 6D pose estimation.

\subsection{Ablation study}
\label{subsection_ablation_study}
To demonstrate the importance of our proposed 6D augmentation, described in \autoref{subsection_6d_augmentation}, we trained a $\phi = 0$ model with and without the 6D augmentation. To gain further insights into the influence of the rotation and scaling part respectively, we also performed experiments in which only one part of the augmentation is used. The color space augmentation is applied in all the experiments to isolate the effect of the 6D augmentation. Due to computational constraints, we performed these experiments only on the driller object from Linemod.\\
\begin{table}
\begin{tabularx}{\columnwidth}{ | r | X  X  X  X |}
 \hline
 Method & w/o 6D & w/ 6D & only scale & only rotation \\
 \hline
 driller & 72.15 & \textbf{99.90} & 97.13 & 97.92 \\
 \hline
\end{tabularx}
\caption{Ablation study to evaluate the influence of our proposed 6D augmentation and it's individual parts. The reported results are in terms of the ADD(-S) metric and are obtained using our $\phi = 0$ model, trained on the driller object of the Linemod dataset.}
\label{table_6d_augmentation_analysis}
\end{table}
 As can be seen from the results in \autoref{table_6d_augmentation_analysis}, the 6D augmentation is a key element in our approach and boosts the performance significantly from 72.15\% without 6D augmentation to 99.9\% in terms of ADD metric. Furthermore, the results from the experiments using only one part of the 6D augmentation (only scale or only rotation) show very similar improvements which suggests that they contribute equally to the overall effectiveness of the 6D augmentation.

\section{Conclusion}
\label{section_conclusion}
In this paper we introduce EfficientPose, a highly scalable end-to-end 6D object pose estimation approach that is based on the state-of-the-art 2D object detection architecture family EfficientDet\cite{EfficientDet}. We extend the architecture in an intuitive and efficient way to maintain the advantages of the base network and to keep the additional computational costs low while performing not only 2D object detection but also 6D object pose estimation of multiple objects and instances - all within a single shot. Our approach achieves a new state-of-the-art result on the widely-used benchmark dataset Linemod while still running end-to-end at over 27 FPS. We thus state that holistic approaches for direct 6D object pose estimation can compete in terms of accuracy with 2D+PnP methods under similar training data conditions - a gap that we close with our proposed 6D augmentation. Moreover, in contrast to 2D+PnP approaches, the runtime of our method is also nearly independent from the number of objects which makes it suitable for real world scenarios like robotic grasping or autonomous driving, where multiple objects are involved and real-time constraints are given.

\bibliographystyle{style/ieee_fullname}
\bibliography{literatur}

\end{document}